\documentclass[11pt]{article}

\usepackage[utf8]{inputenc} % allow utf-8 input
\usepackage[T1]{fontenc}    % use 8-bit T1 fonts
\usepackage{hyperref}       % hyperlinks
\usepackage{url}            % simple URL typesetting
\usepackage{booktabs}       % professional-quality tables
\usepackage{amsfonts}       % blackboard math symbols
\usepackage{textcomp}
\usepackage{csquotes}

\usepackage[a4paper, left=3cm, right=2.5cm, top=3cm, bottom=3cm]{geometry}

\usepackage{authblk}
\usepackage{natbib}

\usepackage{algorithm}
\usepackage{algpseudocode}

\usepackage{graphicx}
\usepackage{tango-colors}
\usepackage{amsmath}
\usepackage{amssymb}
\usepackage{amsfonts}
\usepackage{bm}

\usepackage{booktabs}
\usepackage{multirow}

\usepackage{pgfplots}
\pgfplotsset{compat=1.16}
\usepgfplotslibrary{groupplots,fillbetween}

\usepackage{tikz}
\usetikzlibrary{patterns}

\newcommand{\R}{\mathbb{R}}
\newcommand{\descx}{\mathcal{X}}
\newcommand{\descy}{\mathcal{Y}}

\renewcommand{\cite}{\citep}

\tikzstyle{vec}=[rounded corners, rectangle, semithick, minimum width=1.5cm]
\tikzstyle{inp}=[fill=skyblue1, draw=skyblue3, text=skyblue3]
\tikzstyle{stk}=[fill=orange1, draw=orange3, text=orange3]
\tikzstyle{nrn}=[fill=aluminium4, draw=aluminium6, text=aluminium6]

\tikzstyle{curves}=[line width=0.5mm]
\tikzstyle{curves_thin}=[line width=0.25mm]
\tikzstyle{curves_thick}=[line width=0.75mm]
\tikzstyle{class0color}=[aluminium6]
\tikzstyle{class0}=[draw=aluminium6, fill=aluminium4, text=aluminium6]
\tikzstyle{class1color}=[skyblue3]
\tikzstyle{class1}=[draw=skyblue3, fill=skyblue1, text=skyblue3]
\tikzstyle{class2color}=[orange3]
\tikzstyle{class2}=[draw=orange3, fill=orange1, text=orange3]
\tikzstyle{class3color}=[chameleon3]
\tikzstyle{class3}=[draw=chameleon3, fill=orange1, text=chameleon3]

\begin{document}
	
\title{An A*-algorithm for the Unordered Tree Edit Distance with Custom Costs}

\author[1]{Benjamin Paaßen%
\thanks{Funding by the German Research Foundation (DFG) under grant number PA 3460/2-1 is gratefully acknowledged.}}
\affil[1]{Humboldt-University of Berlin}

\date{This is a preprint version of \citet{Paassen2021SISAP} as provided by the authors. For the original version, refer to Springer.} 
\pagestyle{myheadings}
\markright{This is a preprint of \citet{Paassen2021SISAP} as provided by the authors.}

\maketitle
\title

\begin{abstract}
The unordered tree edit distance is a natural metric to compute distances between trees without 
intrinsic child order, such as representations of chemical molecules.
While the unordered tree edit distance is MAX SNP-hard in principle, it is feasible for small cases, 
e.g.\ via an A* algorithm.
Unfortunately, current heuristics for the A* algorithm assume unit costs for deletions,
insertions, and replacements, which limits our ability to inject domain knowledge.
In this paper, we present three novel heuristics for the A* algorithm that work with
custom cost functions. In experiments on two chemical data sets, we show that custom costs
make the A* computation faster and improve the error of a $5$-nearest neighbor regressor,
predicting chemical properties. We also show that, on these data, polynomial edit distances
can achieve similar results as the unordered tree edit distance.

\end{abstract}

\section{Introduction}

Tree structures occur whenever data follows a hierarchy or a branching pattern, like in
chemical molecules \cite{Gallicchio2013,Rarey1998}, in RNA secondary structures \cite{Shapiro1990},
or in computer programs \cite{Paassen2018ICML}. To perform similarity search on such data,
we require a measure of distance over trees. A popular choice is the tree edit distance,
which is defined as the cost of the cheapest sequence of deletions, insertions, and relabelings 
that transforms one tree to another \cite{Bille2005,Zhang1996,Zhang1989}.
Unfortunately, the edit distance becomes MAX SNP-hard for unordered trees, like tree representations
of chemical molecules \cite{Zhang1994}. Still, for smaller trees, we can compute the unordered tree 
edit distance (UTED) exactly using strategies like A* algorithms~\cite{Horesh2006,Yoshino2013}. 
Roughly speaking, an A* algorithm starts with an empty edit sequence and then successively
extends the edit distance such that a heuristic lower bound for the cost of the edit sequence
remains as low as possible.
The tighter our lower bound $h$, the more we can prune the search and the faster the A* algorithm
becomes. \citet{Horesh2006} have proposed a heuristic based on the histogram of node degrees
and  \citet{Yoshino2013} have improved upon this heuristic by also considering label histograms
and by re-using intermediate values via dynamic programming.
However, both approaches assume unit costs, i.e.\ that deletions, insertions, and relabelings all
have a cost of $1$, irrespective of the content that gets deleted, inserted, or relabeled.
This is unfortunate because, in many domains, we have prior knowledge that suggests different
costs or we may wish to learn costs from data
\cite{Paassen2018ICML}. 
Accordingly, most tree edit distance algorithms are general enough to support custom deletion,
insertion, and replacement costs, as long as they conform to metric constraints \cite{Bille2005,Zhang1996,Zhang1989}.

In this paper, we develop three novel heuristics for the A* algorithm which all support custom
costs. The three heuristics have linear, quadratic, and cubic complexity, respectively,
where the slower heuristics provide tighter lower bounds. Based on these novel heuristics,
we investigate three research questions:
\begin{description}
\item[RQ1:] Which of the novel heuristic is the fastest? And how do they compare against
the state-of-the-art by \citet{Yoshino2013}?
\item[RQ2:] Do custom edit costs actually contribute to similarity search?
\item[RQ3:] How does UTED compare to polynomial edit distances in similarity search?
\end{description}

We investigate these research questions on two example data sets of chemical molecules, both
represented as unordered trees. To answer RQ2 and RQ3, we consider a regression task where we try
to predict the chemical properties of a molecule (boiling point and stability, respectively) via
a nearest-neighbor regression.
We begin our paper with more background and related work before we describe our proposed
A* algorithm, present our experiments, and conclude.

\section{Background and Related Work}

Let $\Sigma$ be an arbitrary set which we call \emph{alphabet}. Then,
we define a \emph{tree} over $\Sigma$ as an expression of the form $\hat x =
x(\hat y_1, \ldots, \hat y_K)$, where $x \in \Sigma$ and where $\hat y_1, \ldots, \hat y_K$
is a list of trees over $\Sigma$, which we call the \emph{children} of $\hat x$.
If $K = 0$, we call $x()$ a \emph{leaf}. We denote the set of all trees over $\Sigma$
as $\mathcal{T}(\Sigma)$.

In this paper, we are concerned with similarity search on trees. In the literature, there are
three general strategies to compute similarities on trees. First, we can construct a feature mapping
$\phi : \mathcal{T}(\Sigma) \to \R^n$, which maps an input tree to a feature vector, and then
compute a (dis-)similarity between features, e.g.\ via $d(\hat x, \hat y) = \lVert \phi(\hat x) - \phi(\hat y)\rVert$.
For example, we can represent trees by $pq$-grams \cite{Augsten2008}, by counts of typical
tree patterns \cite{Collins2002}, or by training a neural network \cite{Gallicchio2013,Kusner2017}.
The second strategy are tree kernels $k$, i.e.\ functions that directly
compute inner products $k(\hat x, \hat y) = \phi(\hat x)^T \cdot \phi(\hat y)$ without the need to
explicitly compute $\phi$ \cite{Aiolli2015,Collins2002}.

\newcommand{\dfsindex}[3]{
\node[#3 of=#1, node distance=0.25cm, orange3] {\scriptsize #2};
}

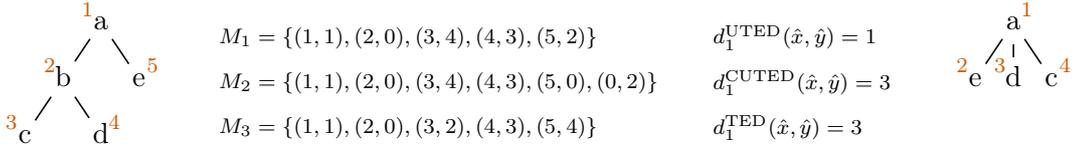
\begin{figure}
\begin{center}
\begin{tikzpicture}[outer sep=0cm, inner sep=0.05cm]
\begin{scope}[shift={(-6,0)}]
\node (xa) at (0,0) {$\strut$a};
\node (xb) at (-0.5,-0.75) {$\strut$b};
\node (xc) at (-1,-1.5) {$\strut$c};
\node (xd) at (0,-1.5) {$\strut$d};
\node (xe) at (+0.5,-0.75) {$\strut$e};

\dfsindex{xa}{1}{above left}
\dfsindex{xb}{2}{above left}
\dfsindex{xc}{3}{above left}
\dfsindex{xd}{4}{above right}
\dfsindex{xe}{5}{above right}

\path[-, semithick]
(xa) edge (xb) edge (xe)
(xb) edge (xc) edge (xd);
\end{scope}

\begin{scope}[shift={(+6,0)}]
\node (ya) at (0,0) {$\strut$a};
\node (ye) at (-0.5,-0.75) {$\strut$e};
\node (yd) at (0,-0.75) {$\strut$d};
\node (yc) at (+0.5,-0.75) {$\strut$c};

\dfsindex{ya}{1}{above right}
\dfsindex{ye}{2}{above left}
\dfsindex{yd}{3}{above left}
\dfsindex{yc}{4}{above right}

\path[-, semithick]
(ya) edge (yc) edge (yd) edge (ye);
\end{scope}

\begin{scope}[shift={(-4.5,0)}]
\node[right] at (0,-0.2) {\scriptsize $M_1 = \{(1,1), (2,0), (3,4), (4,3), (5, 2)\}$};
\node[right] at (0,-0.8) {\scriptsize $M_2 = \{(1,1), (2,0), (3,4), (4,3), (5,0), (0,2)\}$};
\node[right] at (0,-1.4) {\scriptsize $M_3 = \{(1,1), (2,0), (3,2), (4,3), (5,4)\}$};
\end{scope}

\begin{scope}[shift={(2,0)}]
\node[right] at (0,-0.2) {\scriptsize $d_1^\text{UTED}(\hat x, \hat y) = 1$};
\node[right] at (0,-0.8) {\scriptsize $d_1^\text{CUTED}(\hat x, \hat y) = 3$};
\node[right] at (0,-1.4) {\scriptsize $d_1^\text{TED}(\hat x, \hat y) = 3$};
\end{scope}

\end{tikzpicture}
\end{center}
\vspace{-0.6cm}
\caption{An illustration of mappings according to the unordered tree edit distance 
\cite{Zhang1994} (top), the constrained unordered tree edit distance \cite{Zhang1996} (center),
and the ordered tree edit distance \cite{Zhang1989} (bottom) between the same two trees.
The distances assume unit costs.
Numbers in superscript show the depth first order.}
\label{fig:edits}
\end{figure}

In this paper, we focus on a third option, namely tree edit distances~\cite{Bille2005}.
Let $\Sigma$ be an alphabet with $- \notin \Sigma$.
Roughly speaking, a tree edit distance $d(\hat x, \hat y)$ between two trees $\hat x$ and $\hat y$
from $\mathcal{T}(\Sigma)$
is the cost of the cheapest sequence of deletions, insertions, and relabelings of nodes
in $\hat x$ such that we obtain $\hat y$ \cite{Bille2005,Zhang1996,Zhang1989}. 
More precisely, let $x_1, \ldots, x_m$ be the nodes of $\hat x$\footnote{%
Note that we use 'node' and 'label' interchangeably in this paper.
To disambiguate between two nodes with the same label, we use the index $i$.}
and $y_1, \ldots, y_n$ be the nodes of $\hat y$ in depth-first-search order.
Then, we define a \emph{mapping} between $\hat x$ and $\hat y$ as a set of tuples
$M \subset \{0, 1, \ldots, m\} \times \{0, 1, \ldots, n\}$ such that each $i \in \{1, \ldots, m\}$
occurs exactly once on the left and each $j \in \{1, \ldots, n\}$ occurs exactly once on the right.
Figure~\ref{fig:edits} illustrates three example mappings between the trees
$a(b(c, d), e)$ (left) and $a(e,d,c)$ (right), namely $M_1$, $M_2$, and $M_3$ (center left).
Each mapping $M$ can be translated into a sequence of edits by deleting all nodes
$x_i$ where $(i, 0) \in M$, by replacing nodes $x_i$ with $y_j$ where $(i, j) \in M$ and
$x_i \neq y_j$, and by inserting all nodes $y_j$ where $(0, j) \in M$.
We denote the set of all mappings between $\hat x$ and $\hat y$ as $\mathcal{M}(\hat x, \hat y)$.
Next, we define a \emph{cost function} as
a metric  $c : (\Sigma \cup \{-\}) \times (\Sigma \cup \{-\}) \to \R$, and we define the
cost of a mapping $M$ as $c(M) = \sum_{(i, j) \in M} c(x_i, y_j)$ where $x_0 = y_0 = -$.
A typical cost function is $c_1(x, y) = 1$ if $x \neq y$ and $c_1(x, y) = 0$ if $x = y$, which
we call \emph{unit costs}.
Finally, we define the tree edit distance
$d_c : \mathcal{T}(\Sigma) \times \mathcal{T}(\Sigma) \to \R$ according to
cost function $c$ as the minimum $d_c(\hat x, \hat y) = \min_{M \in \mathcal{M}(\hat x, \hat y)} 
c(M)$.

We obtain different edit distances depending on the additional restrictions we apply on the
set of mappings $\mathcal{M}(\hat x, \hat y)$. The unordered tree edit distance (UTED) requires that
mappings respect the ancestral ordering, i.e.\ if $(i, j) \in M$, then descendants of $i$
can only be mapped to descendants of $j$ \cite{Bille2005}. A cheapest example mapping
according to unit costs $M_1$ (Figure~\ref{fig:edits}, top).
The constrained unordered tree edit distance (CUTED) \cite{Zhang1996} additionally requires
that a deletion/insertion of a node implies either deleting/inserting all of its siblings
or all of its children but one. This forbids $M_1$ and $M_3$, where $b$ is deleted but both
its sibling and more than one child are maintained. $M_2$ is a cheapest mapping according
to CUTED with unit costs.
The ordered tree edit distance (TED) \cite{Zhang1989} requires that the ancestral
ordering and the depth-first ordering is maintained.
Accordingly, neither $M_1$ nor $M_2$ are permitted because they swap the order of $c$ and $d$ around.
$M_3$is a cheapest mapping according to TED with unit costs.
Note that UTED is MAX-SNP hard. However, CUTED and TED are both polynomial \cite{Zhang1996,Zhang1989}
via dynamic programming and we consider them as baselines in our experiments.

\section{Method}

In this section, we explain our proposed A* algorithm for the unordered tree edit distance (UTED).
We begin with the general scheme, which we adapt from \citet{Yoshino2013}, and then
introduce three heuristics to plug into the A* algorithm.

\begin{algorithm}
\caption{The A* algorithm to compute the unordered tree edit distance $d^\text{UTED}_c(\hat x, \hat y)$
between two trees $\hat x$ and $\hat y$, depending on a cost function $c$ and a heuristic $h$.}
\label{alg:astar}
\begin{algorithmic}[1]
\Function{astar\_uted}{trees $\hat x$ and $\hat y$, cost function $c$, heuristic $h$}
%\State Let $x_1, \ldots, x_m$ be the nodes of $\hat x$ in depth-first-search order.
%\State Let $y_1, \ldots, y_n$ be the nodes of $\hat y$ in depth-first-search order.
\State Initialize a priority queue $Q$ with the partial mapping $M = \{(1, 1)\}$
\State $\quad$ and value $c(x_1, y_1) + h(\{2, \ldots, m\}, \{2, \ldots, n\})$.
\While{$Q$ is not empty}
\State Poll partial mapping $M$ with lowest value $f$ from $Q$.
\State $i \gets \min \{ 1, \ldots, m+1\} \setminus I_M$.
\If{$i = m+1$}
\State \Return $c(M \cup \{(0, j) | 1 \leq j \leq n, j \notin J_M\})$.
\EndIf
\State Retrieve $(k, l) \in M$ with largest $k$ such that $x_k$ is ancestor of $x_i$ and $l > 0$.
\State $h^p \gets h\big(\{1, \ldots, m\} \setminus (\descx_k \cup I_M), \{1, \ldots, m\} \setminus (\descy_l \cup J_M)\big)$.
\State $M_0 \gets M \cup \{(i, 0)\}$
\State $h_0 \gets h\big(\descx_k \setminus I_{M_0}, \descy_l \setminus J_{M_0}\big) + h^p$.
\For{$j \in \descy_l \setminus J_M$}
\State Let $y'_0, \ldots, y'_t$ be the path from $y_l$ to $y_j$ in $\hat y$ with $y'_0 = y_l$
and $y'_t = y_j$.
\State $M_j \gets M \cup \{(i, j), (0, y'_1), \ldots, (0, y'_{t-1})\}$.
\State $h_j \gets h\big(\descx_i\setminus I_{M_j},\descy_j\setminus J_{M_j}\big) + h\big(\descx_k \setminus (\descx_i \cup I_{M_j}), \descy_l \setminus (\descy_j \cup J_{M_j})\big) + h^p$.
\EndFor
\State Put $M_j$ with value $c(M_j) + h_j$ onto $Q$ for all $j \in \{0\} \cup (\descy_l \setminus J_M)$.
\EndWhile
\EndFunction
\end{algorithmic}
\end{algorithm}

\paragraph{A* algorithm:} We first introduce a few auxiliary concepts that we require for
the A* algorithm. First, let $M$ be some subset of
$\{0, \ldots, m\} \times \{0, \ldots, n\}$.
Then, we denote with $I_M$ the set $\{i > 0 | \exists j : (i, j) \in M\}$ and with $J_M$ the
set $\{j > 0 | \exists i : (i, j) \in M\}$, i.e.\ the set of left-hand-side and right-hand-side
indices of $M$. Next, let $\hat x$ and $\hat y$ be trees with nodes
$x_1, \ldots, x_m$ and $y_1, \ldots, y_n$, respectively. Then, we define $\mathcal{X}_i$
and $\mathcal{Y}_j$ as the index sets of all descendants of $x_i$ and $y_j$, respectively.
Finally, let $c$ be a cost function. Then, we define a \emph{heuristic} as a function
$h : \mathcal{P}(\{1, \ldots, m\}) \times \mathcal{P}(\{1, \ldots, n\}) \to \R$,
such that for any $I \subseteq \{1, \ldots, m\}$ and $J \subseteq \{1, \ldots, n\}$
it holds
\begin{equation}
h(I,J) \leq \min_{M \in \mathcal{M}^\text{UTED}(\hat x, \hat y)} \sum_{(i, j) \in M : i \in I, j\in J} c(x_i, y_j). \label{eq:heuristic}
\end{equation}

Algorithm~\ref{alg:astar} shows the pseudocode for the A* algorithm.
We initialize a partial mapping $M = \{(1, 1)\}$ which maps the root of $\hat x$ to the root
of $\hat y$. If this is undesired, input trees must first be augmented with a placeholder
root node. Next, we initialize a priority queue $Q$ with $M$ and its lower bound.
Now, we enter the main loop. In each step, we consider the current
partial mapping $M$ with the lowest lower bound $f$ (line 5).
If $I_M$ already covers all nodes in $\hat x$,
we complete $M$ by inserting all remaining nodes of $\hat y$ and return the cost of the resulting
mapping (lines 7-9)\footnote{Strictly speaking, this is only valid if the lower bound $f$ is exact for insertions. This is the case for all heuristics considered in this paper.}.
Otherwise, we extend $M$ by mapping the smallest non-mapped index $i$ either to zero
(lines 12-13), or to $j$ for some available node $y_j$ from $\hat y$ (lines 14-18).
In the latter case, we need to maintain the ancestral ordering of the tree
$\hat y$. Accordingly, we first retrieve the lowest ancestor $x_k$ of $x_i$
such that $(k, l) \in M$ and only permit $i$ to be mapped to descendants $\descy_l$.
Note that $(k, l)$ must exist because we initialized $M$ with $\{(1, 1)\}$.
Further, if we map $i$ to a non-direct descendant of $y_l$, we make sure to
insert all nodes on the ancestral path $y'_0, \ldots, y'_t$, first. We generate lower bounds
$h_j$ for all extensions $M_j$ and put them back onto the priority queue.

Note that the space complexity of this algorithm can be polynomially limited by representing
the partial mappings in a tree structure. However, the worst-case time complexity remains
exponential because the algorithm may need to explore combinatorially many possible mappings.
Generally, though, the tighter the lower bound provided by $h$, the fewer partial mappings need
to be explored before we find a complete mapping. To further cut down the time complexity, we  
tabulate the lower bounds $h^p$ for the ancestor mappings $(k, l)$
(line 11), as recommended by \citet{Yoshino2013}.

\paragraph{Heuristics:} The final ingredient we need is the actual heuristic $h$.
We define three heuristics in increasing relaxation and decreasing time complexity.
First,
\begin{equation}
h_3(I, J) = \min_{M \subseteq \mathcal{M}(I, J)} \sum_{(i, j) \in M} c(x_i, y_j),
\end{equation}
where $\mathcal{M}(I, J)$ denotes the set of all mappings between $I$ and $J$, irrespective of
ancestral ordering. Accordingly, Inequality~\ref{eq:heuristic} is trivially
fulfilled because any mapping that respects ancestral ordering is also in $\mathcal{M}(I, J)$.
Importantly, this relaxation can be solved in $\mathcal{O}((m+n)^3)$ via the Hungarian
algorithm \cite{Bougleux2017}. While polynomial, this appears rather expensive for a heuristic.
Therefore, we also consider further relaxations. Without loss of generality, let $|I| \geq |J|$,
otherwise exchange the roles of $\hat x$, $\hat y$, $I$ and $J$. Then, we define
\begin{equation}
h_2(I, J) = \min_{I' \subseteq I : |I'| = |I| - |J|} \Big(\sum_{i \in I'} c(x_i, -)\Big) + \Big(\sum_{i \in I \setminus I'} \min_{j \in J} c(x_i, y_j)\Big). \label{eq:h2}
\end{equation}
Note that this is a lower bound for $h_3(I, J)$ because we expand the class of permitted $M$ to
one-to-many mappings, which is a proper superset of $\mathcal{M}(I, J)$. Further, $h_2$ can be solved in
$\mathcal{O}(m \cdot n)$ because we can evaluate $c_i := \min_{j \in J} c(x_i, x_j)$ for all
$i$ in $|I| \cdot |J|$ steps and we can solve the outer minimization by finding the
$|I| - |J|$ smallest terms according to $c(x_i, -) - c_i$ and using those as $I'$, which is possible
in $O(|I|)$.
In case even $\mathcal{O}(m \cdot n)$ is too expensive, we relax further to
\begin{equation}
h_1(I, J) = \min_{I' \subseteq I : |I'| = |I| - |J|} \sum_{i \in I'} c(x_i, -).
\end{equation}
This is obviously a lower bound for $h_2$ and can be solved in $\mathcal{O}(\max\{m, n\})$.

%In summary, we introduced three heuristics $h_1$, $h_2$, and $h_3$, such that $h_1(I, J)$ $\leq h_2(I, J)
%\leq h_3(I, J)$ for any two sets $I$ and $J$, and which can be computed in linear, quadratic, and
%cubic time, respectively.
%In our experimental evaluation, we will check how the trade-off of tighter lower bounds versus
%time complexity of the heuristic influences the runtime in practice.

\section{Experiments}

We evaluate our three research questions on two data sets from Chemistry, namely the Alkanes data set
of 150 alkane molecules by \citet{Gallicchio2013} and the hundred smallest molecules
from the ZINC molecule data set of \citet{Kusner2017}. In the former case, the molecules
are directly represented as trees (with $8.87$ nodes on average) with hydrogen counts as node labels.
In the latter case, we use the syntax tree of the molecule's SMILES representation (with $13.82$ nodes on average) \cite{Weininger1988}, where nodes are labeled with syntactic blocks. Note that this is a lossy representation because we cut
aromatic rings to obtain trees.

Regarding RQ1, we compute all pairwise UTED values using the three heuristics $h_1$, $h_2$, and $h_3$,
both with unit costs and with custom costs. As custom cost function $c$, we use the difference in 
hydrogen count between two carbon for the alkanes data set. For the ZINC data set, we use the
difference in electron count. For further reference, we also compare to the heuristic of \citet{Yoshino2013} for unit costs. We execute all computations in Python on a consumer desktop PC with Intel core
i9-10900 CPU and 32 GB RAM and measure time using Python's \texttt{time} function.
All experimental code is available at \url{https://gitlab.com/bpaassen/uted}.

\begin{table}
\caption{The average runtime in milliseconds (top) and the number of partial mappings searched
(bottom) per distance computation for each heuristic.}
\label{tab:runtime}
\vspace{-0.5cm}
\begin{center}
\begin{tabular}{ccccccccc}
&& \multicolumn{4}{c}{unit} & \multicolumn{3}{c}{custom} \\
& data set & $h_{\text{1}}$ & $h_{\text{2}}$ & $h_{\text{3}}$ & $h_{\text{yoshino}}$ & $h_{\text{1}}$ & $h_{\text{2}}$ & $h_{\text{3}}$ \\
\cmidrule(lr){1-1} \cmidrule{2-2}  \cmidrule(lr){3-6} \cmidrule(lr){7-9}
\multirow{2}{*}{runtime} & alkanes & $8.70$ & $12.15$ & $10.72$ & $9.52$ & $\bm{7.34}$ & $8.21$ & $9.92$ \\
& ZINC & $549.38$ & $277.15$ & $192.97$ & $266.66$ & $130.62$ & $75.53$ & $\bm{68.12}$ \\
\cmidrule(lr){1-1} \cmidrule{2-2}  \cmidrule(lr){3-6} \cmidrule(lr){7-9}
\multirow{2}{*}{search size} & alkanes & $376$ & $348$ & $260$ & $279$ & $318$ & $302$ & $\bm{246}$ \\
& ZINC & $24586$ & $9164$ & $4158$ & $6781$ & $6643$ & $2655$ & $\bm{1379}$ \\
\end{tabular}
\end{center}
\end{table}

\begin{figure}
\begin{tikzpicture}
\begin{groupplot}[group style={group size=2 by 1,
%	x descriptions at=edge bottom,y descriptions at=edge left,
horizontal sep=1.2cm, vertical sep=0.2cm},
width=6cm,height=4.5cm,xlabel={$m\cdot n$}]
\nextgroupplot[title={alkanes},xmode=log,ymode=log,ylabel={time [s]}]
% background shadings
\addplot[name path=time3custom_lo,draw=none] table[x=x,y=time_lo_3_custom] {time_regression_alkanes.csv};
\addplot[name path=time3custom_hi,draw=none] table[x=x,y=time_hi_3_custom] {time_regression_alkanes.csv};
\addplot[pattern=crosshatch, pattern color=plum3, opacity=0.7] fill between[of=time3custom_lo and time3custom_hi];
\addplot[name path=timeyoshino_lo,draw=none] table[x=x,y=time_lo_yoshino_unit] {time_regression_alkanes.csv};
\addplot[name path=timeyoshino_hi,draw=none] table[x=x,y=time_hi_yoshino_unit] {time_regression_alkanes.csv};
\addplot[orange3, opacity=0.5] fill between[of=timeyoshino_lo and timeyoshino_hi];
\addplot[skyblue1] table[x=x,y=time_1_unit] {time_regression_alkanes.csv};
\addplot[skyblue2] table[x=x,y=time_2_unit] {time_regression_alkanes.csv};
\addplot[skyblue3] table[x=x,y=time_3_unit] {time_regression_alkanes.csv};
\addplot[orange3] table[x=x,y=time_yoshino_unit] {time_regression_alkanes.csv};
\addplot[plum1, densely dashed, thick] table[x=x,y=time_1_custom] {time_regression_alkanes.csv};
\addplot[plum2, densely dashed, thick] table[x=x,y=time_2_custom] {time_regression_alkanes.csv};
\addplot[plum3, densely dashed, thick] table[x=x,y=time_3_custom] {time_regression_alkanes.csv};
\nextgroupplot[title={ZINC},xmode=log,ymode=log,legend pos=outer north east, legend cell align=left]
% background shadings
\addplot[forget plot, name path=time3custom_lo,draw=none] table[x=x,y=time_lo_3_custom] {time_regression_ZINC.csv};
\addplot[forget plot, name path=time3custom_hi,draw=none] table[x=x,y=time_hi_3_custom] {time_regression_ZINC.csv};
\addplot[forget plot, pattern=crosshatch, pattern color=plum3, opacity=0.6] fill between[of=time3custom_lo and time3custom_hi];
\addplot[forget plot, name path=timeyoshino_lo,draw=none] table[x=x,y=time_lo_yoshino_unit] {time_regression_ZINC.csv};
\addplot[forget plot, name path=timeyoshino_hi,draw=none] table[x=x,y=time_hi_yoshino_unit] {time_regression_ZINC.csv};
\addplot[forget plot, orange3, opacity=0.5] fill between[of=timeyoshino_lo and timeyoshino_hi];
\addplot[skyblue1] table[x=x,y=time_1_unit] {time_regression_ZINC.csv};
\addlegendentry{$h_1$, unit}
\addplot[skyblue2] table[x=x,y=time_2_unit] {time_regression_ZINC.csv};
\addlegendentry{$h_2$, unit}
\addplot[skyblue3] table[x=x,y=time_3_unit] {time_regression_ZINC.csv};
\addlegendentry{$h_3$, unit}
\addplot[orange3] table[x=x,y=time_yoshino_unit] {time_regression_ZINC.csv};
\addlegendentry{$h_\text{yoshino}$, unit}
\addplot[plum1, densely dashed, thick] table[x=x,y=time_1_custom,y error=time_err_1_custom] {time_regression_ZINC.csv};
\addlegendentry{$h_1$, custom}
\addplot[plum2, densely dashed, thick] table[x=x,y=time_2_custom,y error=time_err_2_custom] {time_regression_ZINC.csv};
\addlegendentry{$h_2$, custom}
\addplot[plum3, densely dashed, thick] table[x=x,y=time_3_custom,y error=time_err_3_custom] {time_regression_ZINC.csv};
\addlegendentry{$h_3$, custom}
\end{groupplot}
\end{tikzpicture}
\vspace{-0.4cm}
\caption{A log-log regression of the runtime needed for computing UTED for all four heuristics
(indicated by color) on the alkanes data (left)
and the ZINC data (right). Shading indicates distance between 25th and 75th percentile of the runtimes for $h_\text{yoshino}$ (orange, solid), and $h_3$ with custom costs (purple, crosshatch),
respectively.}
\label{fig:runtime}
\end{figure}
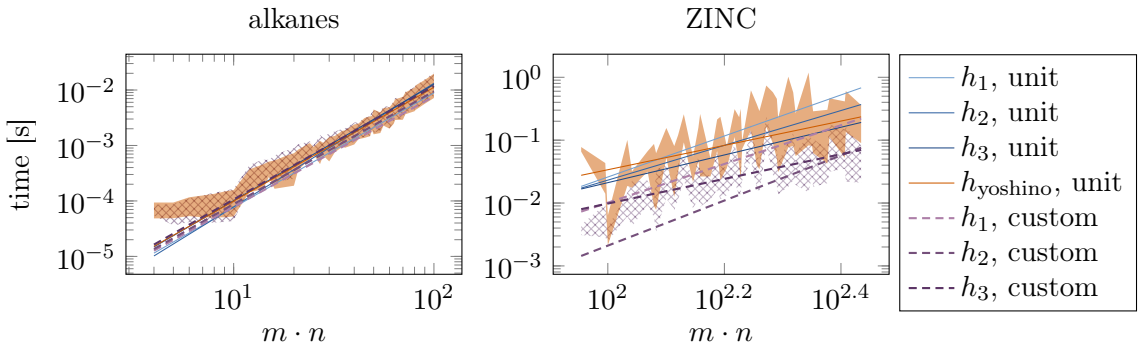

Table~\ref{tab:runtime} shows the average runtime in milliseconds (top) for each heuristic
on both data sets. On alkanes, $h_1$ is fastest and on ZINC, $h_3$ is fastest. All heuristics
get faster for custom costs.
Surprisingly, $h_\text{yoshino}$ is not the fastest for
unit costs, even though it is optimized for this setting. This may just be due
to an unfavourable constant factor, though: $h_\text{yoshino}$ is successful in reducing the
size of the search space almost to the same level as $h_3$ (see Table~\ref{tab:runtime}, bottom).
Further, Figure~\ref{fig:runtime} displays a linear regression for the runtime versus tree size
in a log-log plot, indicating that $h_\text{yoshino}$ and $h_3$ have the lowest slopes/best
scaling behavior for large trees.

\begin{table}
\caption{Average RMSE ($\pm$ std.) of a $5$-NN regressor across $15$ crossvalidation folds
for UTED, CUTED, and TED with unit and custom costs.}
\label{tab:rmse}
\vspace{-0.5cm}
\begin{center}
\begin{tabular}{ccccccc}
& \multicolumn{3}{c}{unit} & \multicolumn{3}{c}{custom} \\
data set & UTED & CUTED & TED & UTED & CUTED & TED \\
\cmidrule(lr){1-1}  \cmidrule(lr){2-4} \cmidrule(lr){5-7}
alkanes & $0.27 \pm 0.24$ & $0.27 \pm 0.24$ & $0.27 \pm 0.24$ & $\bm{0.25 \pm 0.24}$ & $\bm{0.25 \pm 0.24}$ & $\bm{0.25 \pm 0.24}$ \\
ZINC & $1.33 \pm 0.85$ & $1.31 \pm 0.86$ & $1.36 \pm 0.84$ & $\bm{1.24 \pm 0.87}$ & $1.26 \pm 0.87$ & $1.29 \pm 0.86$ \\
\end{tabular}
\end{center}
\end{table}

Regarding RQ2 and RQ3, we perform a $5$-nearest neighbor regression%
\footnote{We also tested lower $K$, which achieved worse results for all methods.}
to predict the boiling point of alkanes and the chemical stability measure of \citet{Kusner2017}
for ZINC molecules, respectively. Table~\ref{tab:rmse} shows the prediction error for
both data sets in 15-fold crossvalidation. For reference, we do not only evaluate UTED with unit and custom costs, but
also CUTED and TED. We observe that all methods perform better with custom costs
compared to unit costs. For alkanes, there is no measurable difference between UTED, CUTED,
and TED. For ZINC, TED performs worst and CUTED performs better than UTED for unit costs
and UTED performs better than CUTED for custom costs.

\section{Conclusion}

We proposed three novel heuristics to compute the unordered tree edit distance via an
A* algorithm. In contrast to prior work, our heuristics can accommodate custom costs,
not only unit costs. Our three heuristics provide different trade-offs of time complexity
(linear, quadratic, cubic) versus how much they prune the A* search.

In our experiments on two chemical experiments, we observed that this trade-off works
in favor of the linear heuristic for small trees but that the cubic heuristic takes
over for larger trees. Interestingly, the cubic heuristic compared favorably
even to the current state-of-the-art heuristic.
When applying custom costs, all our heuristics became faster thanks to the disambiguation
provided by the custom cost function.

Regarding similarity search, we investigated the performance of a $5$-nearest neighbor
regressor, predicting chemical properties. We observed that custom costs lowered the
regression error. However, we also saw that a similar performance can be achieved with
a polynomial, restricted edit distance. Future work might investigate further tree data
set to check whether these results generalize beyond chemistry.

\bibliographystyle{elsarticle-num-names}
\bibliography{literature}

\end{document}